\documentclass[letterpaper, 10 pt, conference]{ieeeconf}  

\IEEEoverridecommandlockouts                              
\usepackage[T1]{fontenc}
\usepackage{censor}
\usepackage{graphicx}
\usepackage{amsfonts}
\usepackage{amsmath}
\usepackage{bm}
\usepackage[hyphens]{url}
\usepackage{xurl}

\usepackage[ruled,vlined]{algorithm2e}
\usepackage{cite}
\usepackage{booktabs}
\usepackage{tabularx,array,textcomp}
\usepackage{multirow}
\usepackage{float}
\usepackage{placeins}

\title{\LARGE \bf
Recursive Self-Improvement of Exoskeleton Assistance Policies through Human-in-the-Loop Feedback
}

\author{Muyuan Ma$^{\dagger}$, Shuo Jiang$^{\dagger}$,
Lijun Han, Xinpan Meng, Zhenghua Ma, Xiuze Xia,
Houcheng Li, and Long Cheng$^{*}$%
\thanks{$^{\dagger}$Muyuan Ma and Shuo Jiang contributed equally to this work.}%
\thanks{$^{*}$Corresponding author: Long Cheng.}%
\thanks{The authors are with the School of Artificial Intelligence,
University of Chinese Academy of Sciences, Beijing 100049, China,
and also with the State Key Laboratory of Multimodal Artificial
Intelligence Systems, Institute of Automation, Chinese Academy
of Sciences, Beijing 100190, China
(e-mail: long.cheng@ia.ac.cn).}%
}
\begin{document}

\maketitle
\thispagestyle{empty}
\pagestyle{empty}

\begin{abstract}
To reduce reliance on manual design and extensive parameter tuning, a framework for the autonomous evolution of exoskeleton assistance policies through human feedback is proposed. Assistance policies are autonomously generated as executable control code by an intelligent agent and evaluated in experiments with human participants. Subjective user feedback, movement performance, and muscle activation measurements are incorporated into policy assessment. Based on these assessments, policy deficiencies are autonomously identified and the control code is revised by the agent for subsequent trials. A closed loop of policy generation, human feedback, evaluation, and autonomous refinement is thereby established. The framework is evaluated using GPT, DeepSeek, GLM, and Kimi, and initial assistance performance and refinement outcomes are compared across models. Despite differences in initial performance, effective assistance is achieved with all four models after a single autonomous refinement iteration under the tested conditions. Across three tasks, biceps and triceps muscle activation levels are reduced by 29.4\% and 35.2\%, respectively, compared with those measured without an exoskeleton. However, limited adaptability to unexpected situations is observed, indicating that further improvements in robustness and safety are required.
\end{abstract}

\section{Introduction}   
Wearable exoskeletons have been developed to augment human physical capabilities and reduce the effort required for locomotion and occupational activities. Reductions in metabolic expenditure during walking and running have been demonstrated with portable exosuits~\cite{kim2019exosuit}, while muscular demands during industrial work have been alleviated through wearable shoulder assistance~\cite{zhou2024shoulder}. However, effective assistance cannot be determined solely from actuator capability or mechanical design. Its benefits also depend on how assistance is coordinated with the wearer's movements and adapted to individual responses~\cite{slade2024hil}. When predefined assistance profiles are employed, their timing, magnitude, and other control parameters must be appropriately selected. Consequently, the development of effective assistance remains closely associated with controller design, experimental evaluation, and repeated refinement.

To reduce dependence on manual tuning, human-in-the-loop (HIL) optimization has been introduced to adjust assistance according to measured human responses. In~\cite{zhang2017walking}, ankle assistance profiles were optimized using a covariance matrix adaptation evolution strategy, with metabolic expenditure employed as the optimization objective. Bayesian optimization was subsequently applied to personalize hip assistance delivered by a soft exosuit~\cite{ding2018hip}. The application of HIL optimization has also been extended to hip assistance during stair climbing~\cite{park2025stairs}. These studies established that assistance can be improved through direct interaction with the wearer rather than being determined entirely from predefined biomechanical assumptions. Nevertheless, repeated human evaluations are required, and the observed benefits are influenced by both controller customization and the wearer's adaptation to assistance, as investigated in~\cite{poggensee2021adaptation}.

To improve the efficiency and practicality of this process, alternative methods for estimating assistance outcomes have been investigated. Online musculoskeletal simulation was used in~\cite{gordon2022simulation} to estimate metabolic cost and optimize hip assistance with a reduced reliance on lengthy calorimetry measurements. In~\cite{slade2022realworld}, personalization during everyday walking was enabled through an estimation approach based on wearable sensor measurements. Electromyography (EMG) has also been incorporated into assistance optimization. Human and robot muscle synergies were exploited to optimize EMG based assistive strategies in~\cite{hamaya2019synergies}, while the selection of muscle activity cost functions for ankle assistance was examined in~\cite{han2021cost}. These approaches provide alternative routes for evaluating assistance, although the choice of physiological measurements and their conversion into optimization objectives remain important design decisions.

Beyond physiological objectives, subjective experience has increasingly been incorporated into assistance personalization. The role of individual preference in customized exoskeleton control was examined in~\cite{ingraham2022preference}, and an active learning approach was developed to identify preferred ankle assistance settings through repeated user comparisons~\cite{lee2023preference}. Preference based optimization was further applied to lifting and lowering with a soft back exosuit~\cite{arens2025back}, while direct adjustment of assistance parameters by the wearer was investigated in~\cite{schaefer2026tuning}. Multiple objectives have also been considered: gait outcomes were jointly optimized for dropfoot assistance in~\cite{zhang2025multiobjective}, and biomechanical metrics were combined with preference information in an offline assistive device tuning framework in~\cite{choe2026balancing}. Thus, the incorporation of subjective feedback, or its combination with objective measurements, cannot by itself distinguish a new assistance optimization method. An additional question is how these observations can be translated into appropriate changes to the executable assistance policy.

Adaptability across wearers and activities has been addressed through complementary approaches. Dynamic movement primitives, Bayesian optimization, and a task translation mechanism were integrated in~\cite{chen2024multitask} to personalize assistance across different wearers and tasks. Biological joint moment estimation was used to unify exoskeleton control in~\cite{molinaro2024unified}, and this direction was extended to assistance across diverse activities without separate controllers for each activity in~\cite{molinaro2024taskagnostic}. These advances demonstrate that contemporary assistance methods are not necessarily restricted to a single movement or a fixed torque profile. Nevertheless, the controller implementation, adaptation mechanism, and evaluation criteria must still be established when constructing such systems.

To reduce the burden of specifying exoskeleton behaviors and improve contextual adaptation, large language models (LLMs) have been incorporated into exoskeleton interfaces and control frameworks. Natural language interfaces were developed and evaluated to support voice interaction with occupational exoskeletons~\cite{moreno2024xonli,moreno2024xonliassessment}. In~\cite{chen2025hand}, an incremental learning framework was proposed in which gesture and grasping programs were generated with LLM assistance and incorporated into a local controller to extend its executable action repertoire. Speech and image inputs were further combined to infer motion intentions and generate executable actions in the embodied interaction control framework~\cite{li2025eic}. In~\cite{chen2026semantic}, task semantics were used to configure control modes, impedance parameters, and movement speed, while anomaly detection was incorporated to trigger replanning. These studies demonstrate the potential of language models for instruction interpretation, action generation, and contextual adjustment of assistance. However, successful command execution and contextual adaptation do not necessarily imply reduced muscular effort or improved perceived assistance. A complementary challenge is therefore to autonomously revise executable assistance code according to subjective feedback and muscle activation measurements obtained from successive human trials, and to verify whether these revisions improve assistance.

To investigate this problem, a framework for self-improving exoskeleton assistance policies through human feedback is proposed in this article. Executable assistance code is autonomously generated by an intelligent agent and evaluated through experiments with a human participant. Subjective feedback, movement performance, and muscle activation measurements are then provided for policy assessment. Based on these observations, deficiencies are identified and the control code is revised by the agent for subsequent trials. Policy generation and revision are performed between experimental trials, forming a closed loop between code development and physical evaluation. The framework is evaluated using GPT, DeepSeek, GLM, and Kimi to examine differences in initial assistance and subsequent refinement. Experiments are conducted across three tasks, with biceps and triceps muscle activation assessed relative to the condition without an exoskeleton. Responses to unexpected situations are also examined to identify limitations beyond performance in the nominal tasks. The main contributions are summarized as follows:

\begin{enumerate}
    \item An intelligent agent framework is developed for the autonomous generation and iterative revision of executable exoskeleton assistance policies, reducing reliance on manual control code development.

    \item A feedback driven refinement procedure is established in which subjective user experience, movement performance, and muscle activation measurements are used to guide the agent's assessment and revision of assistance policies.

    \item The framework is evaluated through physical experiments across three tasks and four model families. Initial and refined policies are compared, and limitations in handling unexpected situations are identified to delineate the demonstrated capabilities of the approach.
\end{enumerate}

\section{Methods}
\label{sec:methods}
\subsection{Overall Framework}

\begin{figure*}[t]
\centering
\includegraphics[width=0.98\textwidth]{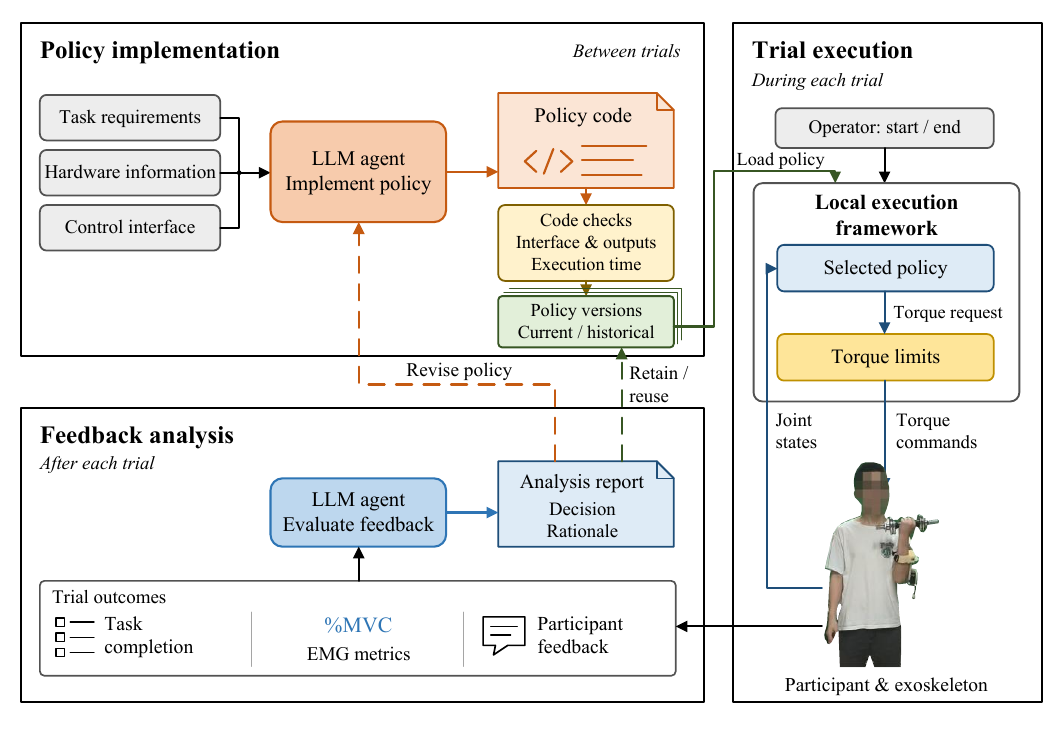}
\caption{An LLM-agent framework for exoskeleton assistance policy generation, execution, and feedback-based refinement. The policy implementation module generates executable code from task requirements, hardware information, and a unified control interface. Candidate code undergoes checks before becoming an available policy version. The local execution framework loads and runs the selected policy during a trial started and ended by the operator. During execution, the framework exchanges joint-state information and torque commands with the exoskeleton. EMG results, task completion, and participant feedback inform policy assessment. The feedback analysis module either proposes a revision based on a selected policy or selects an existing policy for reuse. Code revisions and policy-version selection occur between trials. Solid arrows indicate execution or data flow; dashed arrows indicate revision or reuse decisions.}
\label{fig:framework}
\end{figure*}

We developed an intelligent agent framework based on large language models (LLMs) for generating and iteratively refining exoskeleton assistance policies (Fig.~\ref{fig:framework}). The framework comprises three modules for policy implementation, trial execution, and feedback analysis. The policy implementation module generates executable assistance policies from task requirements, hardware information, and a unified control interface. The trial execution module runs the selected policy during each trial and collects joint motion and electromyography (EMG) data. After each trial, the feedback analysis module evaluates task completion, EMG metrics, and participant feedback to select an existing policy or request a revision. When revision is requested, the policy implementation module updates the policy code between trials for subsequent execution.

The framework aims to reduce the mean activation of each target muscle during the task, with successful task completion as a prerequisite. The policy implementation and feedback analysis modules use LLM-based agents to generate, evaluate, and revise assistance policies. The trial execution module uses a local execution framework running on the host computer to invoke the policy at each control cycle, enforce torque command constraints, and record data. Each trial is started and ended by the operator. After each trial, task completion and participant feedback on perceived assistance and discomfort are recorded.

\subsection{Assistance Policy Generation}

When generating an initial policy from scratch, the policy implementation module produces executable assistance code from task instructions, hardware documentation, and the unified control interface specification. The task instructions define the required movements, loading conditions, enabled sides, and muscles to be evaluated. They direct the agent to reduce each target muscle's mean activation during the task while maintaining successful task completion. The hardware documentation describes the known exoskeleton structure, the mapping between joints and motors, and operating constraints. The control interface specifies the states available to the policy, its output format, and its initialization procedure. At this stage, the agent receives no existing policy code, human EMG results, or historical feedback.

Within the interface and operating constraints, the agent may autonomously design the policy's control logic according to the task requirements. This may include motion-phase estimation, internal states, and filtering. During a trial, the policy computes assistance from observations supplied through the unified interface. Between trials, the agent may revise the control logic and parameters based on feedback analysis.

The agent delivers executable policy code, necessary supporting files, and design documentation. The documentation describes control logic, parameter rationale, initialization, input usage, design assumptions, and anticipated failure scenarios. After delivery, the system executes the code using predefined joint-state sequences to check the input/output interface, numerical outputs, and execution time. Policies that pass these checks become candidates for subsequent human trials.

\subsection{Policy Execution and Runtime Constraints}

During a trial, the local execution framework invokes the generated assistance policy at each control cycle. Available inputs include elbow joint angle, angular velocity, and the previous torque command processed by the framework for each enabled side. The interface also provides elapsed time since trial onset and the interval between consecutive control cycles. Each assistance policy selects inputs according to its control logic and computes a joint torque request for each enabled side. The framework processes these requests, converts them into motor commands, and transmits them to the device. EMG recorded during the trial supports policy evaluation by the feedback analysis module.

At the start of each trial, the local execution framework loads and initializes the selected assistance policy. During the trial, the policy processes real-time inputs and computes assistance torque requests according to its control logic. It may retain and update inferred motion phases, timing information, or intermediate results for use in subsequent control cycles.

For illustration, the policy generated by 6-astra for the dumbbell curl task uses elbow angle, angular velocity, and movement duration to infer motion phases. The policy computes assistance for left elbow flexion based on this phase estimate. Once the duration and angular excursion of flexion reach predefined thresholds, the policy enters the raising phase. Target assistance is based on the sine of elbow angle and modulated by movement velocity and position within the joint range. Assistance is gradually introduced after the start of the raising phase. During normal operation, the target torque for each phase is expressed as

\begin{equation}
\tau_{\mathrm{tar}}(k)=
\begin{cases}
\tau_p\max(0,\sin q_k)\,g_{\mathrm{up}}(k), & \text{raising},\\
B_k, & \text{lowering},\\
0, & \text{idle}.
\end{cases}
\label{eq:target-torque}
\end{equation}

Here, $q_k$ is the elbow joint angle at control cycle $k$. The peak parameter $\tau_p$ is set by the agent and may be adjusted in subsequent policy revisions based on human feedback. The coefficient $g_{\mathrm{up}}(k)$ combines four multiplicative factors: modulation across the joint range, velocity-dependent attenuation of assistance, raising motion, and assistance onset. These factors adjust assistance according to joint position and movement velocity while gradually introducing assistance after the raising phase begins. The updated lowering-phase bound $B_k$ also serves as the target torque during that phase.

When reverse movement or a sustained pause is detected during raising, the policy enters the lowering phase and reduces the target assistance bound. During lowering, this bound is maintained or reduced further according to the movement state, as follows:

\begin{equation}
\begin{aligned}
B_k=\min\bigl\{&B_{\mathrm{pre}}(k),\\
&\alpha\tau_p\max(0,\sin q_k)\,g_{\mathrm{down}}(k)\bigr\}.
\end{aligned}
\label{eq:lowering-bound}
\end{equation}

The coefficient $g_{\mathrm{down}}(k)$ combines three multiplicative factors: modulation across the joint range, velocity-dependent attenuation of assistance, and lowering motion. These factors adjust target assistance according to joint position and lowering velocity. Both $g_{\mathrm{up}}(k)$ and $g_{\mathrm{down}}(k)$ take values within the interval $[0,1]$. The factors for joint-range modulation, velocity-dependent attenuation of assistance, and motion modulation are computed using elbow angle and raw and low-pass-filtered angular velocities. The assistance onset factor in $g_{\mathrm{up}}(k)$ increases smoothly from 0 to 1 over the first 0.65~s of the raising phase. $B_{\mathrm{pre}}(k)$ denotes the bound before its update at the current cycle. Upon each transition into lowering, $B_{\mathrm{pre}}(k)$ is initialized to $\alpha\min\{\tau_{\mathrm{req}}(k-1),\tau_p\}$. In subsequent lowering cycles, it takes the previous bound $B_{k-1}$. Here, $\tau_{\mathrm{req}}(k-1)$ is the previous torque request retained internally by the policy. The illustrated policy uses $\tau_p=1.55\,\mathrm{N\,m}$ and $\alpha=0.45$. The policy uses predefined torque rates to adjust its request $\tau_{\mathrm{req}}(k)$ gradually toward the target torque $\tau_{\mathrm{tar}}(k)$.

The local execution framework uses preset torque magnitude and rate limits. These limits constrain the magnitude of the processed torque command and its change between consecutive control cycles. During normal assistance control cycles, the framework first validates the joint torque request. It then applies magnitude and rate limits sequentially, followed by a final magnitude limit. For each enabled side, the processed joint torque command satisfies the following constraints:

\begin{align}
\lvert\tau_{\mathrm{cmd}}(k)\rvert &\leq L,\label{eq:torque-magnitude}\\
\lvert\tau_{\mathrm{cmd}}(k)-\tau_{\mathrm{cmd}}(k-1)\rvert
&\leq r\Delta t_k.\label{eq:torque-rate}
\end{align}

Here, $\tau_{\mathrm{cmd}}(k)$ denotes the joint torque command after magnitude and rate limiting at cycle $k$. $L$ is the torque magnitude limit ($\mathrm{N\,m}$), and $r$ is the torque rate limit ($\mathrm{N\,m/s}$). $\Delta t_k$ is the interval between consecutive control cycles, expressed in seconds. For the first control calculation, the previous command is initialized as $\tau_{\mathrm{cmd}}(0)=0$. Both $L$ and $r$ remain fixed within each trial, with their values reported in the experimental setup.

During a trial, the local execution framework monitors policy computation time, control-cycle intervals, motor feedback updates, elbow joint angles, and motor status. Detected exceptions include policy execution errors, computation or feedback timeouts, joint-angle limit violations, and motor faults. When an exception is detected, the framework stops subsequent policy calls and attempts to send zero-torque commands to each enabled side. It submits fault records describing the exception and the handling actions for subsequent analysis.

\subsection{Feedback Evaluation and Policy Refinement}
\label{subsec:feedback-refinement}

After each trial, the feedback analysis module evaluates the current assistance policy using EMG results, task completion status, and subjective participant feedback. EMG evaluation uses the mean normalized activation of each target muscle over the task. The raw EMG signals are filtered, and predefined segments at both ends of the task interval are excluded. Root mean square (RMS) values are calculated within complete sliding windows to characterize EMG amplitude~\cite{mcmanus2021terminology}. These values are then normalized to the corresponding muscle's maximum voluntary contraction (MVC) reference~\cite{besomi2020normalization}. The normalized activation within each window and its task mean are defined as

\begin{align}
p_i(j)&=100\,\frac{\mathrm{RMS}_i(j)}{\mathrm{MVC\_RMS}_i},\label{eq:emg-window}\\
\bar p_i&=\frac{1}{K_i}\sum_{j=1}^{K_i}p_i(j).\label{eq:emg-mean}
\end{align}

Here, $\mathrm{RMS}_i(j)$ is the RMS value for muscle $i$ in complete window $j$. $\mathrm{MVC\_RMS}_i$ is the corresponding RMS reference obtained from that muscle's MVC calibration. $K_i$ denotes the number of complete windows included in the task evaluation. The quantities $p_i(j)$ and $\bar p_i$ represent normalized activation within each window and its task mean, respectively, both expressed in \%MVC. The task mean is calculated and reported separately for each target muscle. Section~\ref{sec:experimental-setup} describes the EMG acquisition and processing parameters and the calculation of MVC reference values.

For trials with valid EMG data, each target muscle's task mean $\bar p_i$ is compared with a reference $b_i$ recorded without the exoskeleton. This reference is selected for that trial and denotes the muscle's task mean for the same participant performing the same task without the exoskeleton. Each comparison uses matching EMG processing procedures and the same MVC reference for the trial and its reference measurement. Subtracting $b_i$ from $\bar p_i$ gives the difference in percentage points of \%MVC. For $b_i>0$, the relative change is calculated as follows:

\begin{equation}
\Delta_i=100\,\frac{\bar p_i-b_i}{b_i}.
\label{eq:relative-emg}
\end{equation}

The relative change $\Delta_i$ is expressed as a percentage, with negative and positive values indicating decreased and increased mean normalized activation, respectively. Assistance effectiveness is assessed with successful task completion as a prerequisite, and participant feedback informs the interpretation of EMG changes. For trials in which the task is not completed, available EMG metrics and participant feedback are used to analyze execution difficulties and potential policy problems.

The feedback analysis module combines these evaluations with participant reports of assistance during different movement phases and prior trial records explicitly made available to it. It analyzes policy performance and identifies control elements that may require adjustment. The module may retain the current policy, reuse a previous policy, or select either as the basis for revision. It may propose adjustments to assistance magnitude, assistance timing, phase transitions, or the policy's internal torque rate of change. The module generates an analysis report documenting the selected policy, decision rationale, and supporting evidence. When recommending revision, the module also specifies the proposed adjustments, expected effects, and uncertainties in the report.

When revision is selected, the policy implementation module modifies the control logic or parameters of the selected policy according to the feedback analysis report. It updates both the policy code and the accompanying design documentation. After passing the previously described code checks, the revised policy is saved as a new version, with its changes and base policy recorded. If an existing policy is retained or reused, the selected policy is used in the next trial. Policy code is revised between trials, and the operator starts each subsequent trial. The resulting EMG metrics, task completion status, and participant feedback are used to evaluate the selected policy and inform further adjustments.

\section{Experiments}
\label{sec:experiments}
\subsection{Experimental Setup}
\label{sec:experimental-setup}

\begin{figure*}[t]
\centering
\includegraphics[width=\textwidth]{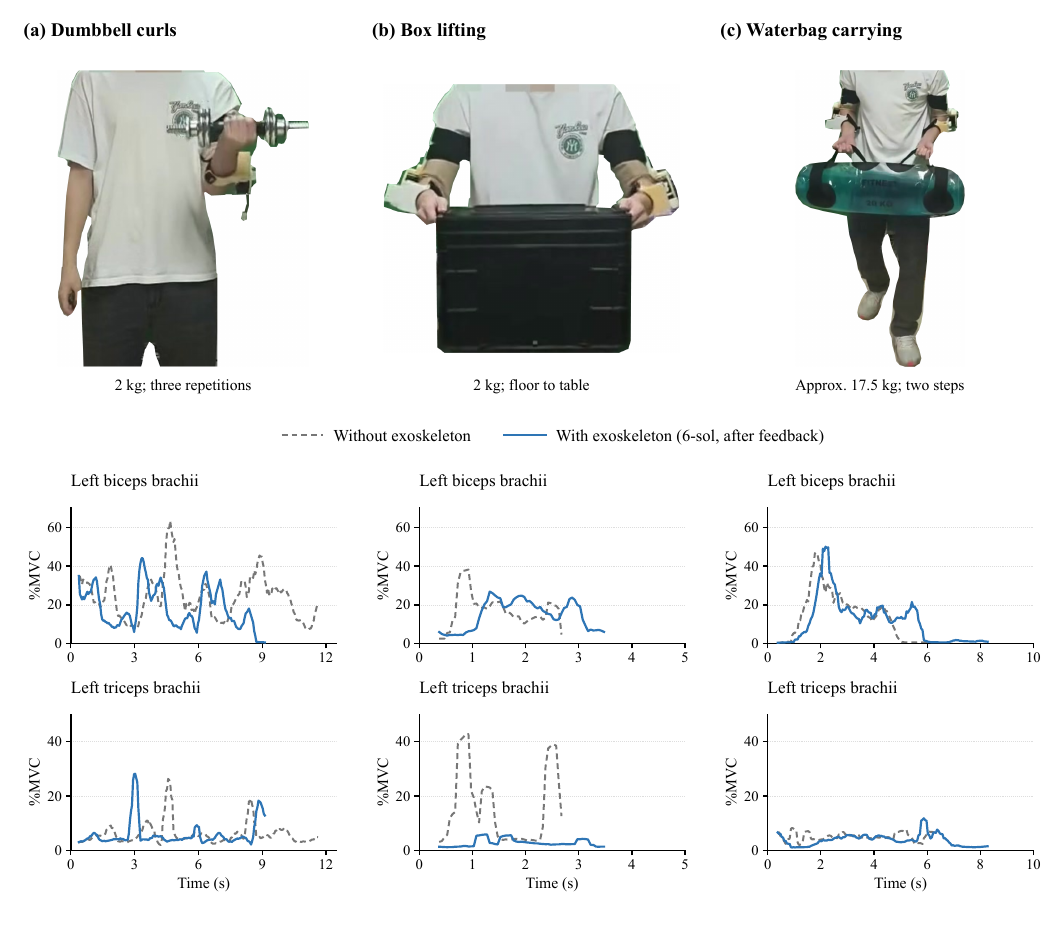}
\caption{Experimental tasks and representative muscle-activation time courses. (a) Three consecutive left-arm curls with a 2-kg dumbbell. (b) Lifting a 2-kg box from the floor onto a table. (c) Lifting an approximately 17.5-kg waterbag, taking two steps, and returning it to the floor. Below each task photograph, curves show MVC-normalized RMS EMG from the left biceps and triceps brachii in one no-exoskeleton trial (gray dashed) and one trial with the 6-sol policy selected after feedback (blue solid). For dumbbell curls, the initial 6-sol policy was retained after feedback. The curves use the same selected records as Fig.~\ref{fig:emg_results}. Each EMG trace retains its original time scale, with zero set independently at the start of its evaluation interval. Photographs illustrate the tasks.}
\label{fig:experimental_setup}
\end{figure*}

We evaluated the assistance policies with one participant performing three tasks: dumbbell curls, box lifting, and waterbag carrying (Fig.~\ref{fig:experimental_setup}). The participant was a 20-year-old, right-handed man (175~cm, 65~kg) who reported no upper-limb injury or disorder. All experimental procedures were reviewed and approved by the ethics committee of the authors' institution. The committee name and approval number are withheld for double-anonymous review and will be disclosed upon acceptance. The participant provided informed consent. The experimental setup comprised the elbow exoskeleton described in~\cite{xia2024compliant}, a host computer, a USB-to-CAN adapter, and an EMG acquisition system. The exoskeleton was actuated by AK60-6 motors at the left and right elbow joints. The host computer ran the local execution framework, sending motor commands and receiving joint-state feedback through the USB-to-CAN adapter. The framework's target control frequency was set to 1000~Hz. During normal assistance control, the framework limited each enabled side's joint torque command to $9\,\mathrm{N\,m}$ in magnitude and $55\,\mathrm{N\,m/s}$ in rate of change.

The Delsys Trigno system was operated through Trigno Control Utility, the manufacturer's software for sensor configuration and data streaming. With upsampling enabled, the system streamed surface EMG signals to the host computer at 2000~Hz. EMG signals were recorded from the biceps brachii and triceps brachii bilaterally during box lifting and waterbag carrying, and only from the left arm during dumbbell curls. The skin at the recording sites was cleaned, and the sensors were secured with tape.

For offline evaluation, the recorded EMG signals were band-pass filtered between 20 and 450~Hz using a Butterworth filter applied in the forward and reverse directions. To reduce filtering edge effects, the first and last 0.25~s of each evaluation interval were excluded. RMS amplitudes were then calculated over complete 0.25-s windows with a 0.05-s step. MVC calibration recordings were processed using the same filtering and RMS procedure. For each muscle, the maximum windowed RMS amplitude across three accepted calibration trials served as the reference for normalizing task EMG. The normalized values were averaged across windows to obtain task mean normalized activation, as defined in Section~\ref{subsec:feedback-refinement}.

\subsection{Tasks and Experimental Protocol}
\label{sec:experimental-protocol}

For dumbbell curls (Fig.~\ref{fig:experimental_setup}(a)), the participant stood holding a 2-kg dumbbell in the left hand, with the upper arm close to the torso and approximately vertical. Each trial comprised three consecutive curls at a self-selected pace, with the arm returning to its initial hanging position after each repetition. Each box-lifting trial (Fig.~\ref{fig:experimental_setup}(b)) involved bending down from standing, grasping a 2-kg box with both hands, and placing it on a table in front of the participant. For waterbag carrying (Fig.~\ref{fig:experimental_setup}(c)), the participant lifted an approximately 17.5-kg waterbag from the floor with both hands until its handles reached approximately waist height. The participant then took two steps while maintaining this height, before returning the bag to the floor and releasing the handles. The box-lifting and waterbag-carrying tasks were each performed once per trial.

We evaluated four model families: GPT (6-astra and 6-sol), DeepSeek (V4.1-Flash), GLM (GLM-5.3), and Kimi (Kimi-K3). Each model was used for both policy implementation and feedback analysis. For each of the three tasks, we evaluated each model's starting policy and the policy selected after the first round of feedback. Assistance performance was assessed using task completion, participant feedback, and mean muscle activation quantified from MVC-normalized EMG (\%MVC). For each target muscle, mean activation during assisted trials was compared with that measured during the same task without the exoskeleton.

The operator manually started and ended each trial, while EMG and joint-state data were recorded during task execution. After each trial, task completion was recorded, and the participant provided feedback on perceived assistance, movement interference, and discomfort. The recorded results and participant feedback were submitted to the corresponding model for the feedback analysis described in Section~\ref{subsec:feedback-refinement}. The policy selected through this process was then tested in a subsequent trial of the same task.

\begin{table*}[!tp]
\centering
\caption{Task completion and participant-reported assistance with initial policies and policies selected after one feedback round.}
\label{tab:assistance_outcomes}
\begingroup
\fontsize{8.3}{9.6}\selectfont
\setlength{\tabcolsep}{3pt}
\renewcommand{\arraystretch}{1.13}
\begin{tabularx}{\textwidth}{@{}p{20mm}cc>{\raggedright\arraybackslash}X>{\raggedright\arraybackslash}X@{}}
\toprule
 & \multicolumn{2}{c}{Task completed} & \multicolumn{2}{c}{Participant feedback} \\
\cmidrule(lr){2-3}\cmidrule(l){4-5}
Model & Initial & \shortstack{After\\feedback} & Initial policy & After feedback \\
\midrule
\multicolumn{5}{@{}l}{\textbf{Dumbbell curls}} \\
\addlinespace[2pt]
6-astra & Yes & Yes & Abrupt, strong lifting assistance; resistance during lowering. & Satisfactory experience; no discomfort. \\
6-sol & Yes & Yes\textsuperscript{\fontsize{6}{6}\selectfont a} & Satisfactory experience; no discomfort. & Satisfactory experience; no discomfort. \\
DeepSeek & No & Yes & Lowering interference prevented three curls. & Highly dissatisfied with assistance. \\
GLM & No & Yes & Lowering interference prevented three curls. & Highly dissatisfied with assistance. \\
Kimi & No & Yes & Lowering interference prevented three curls. & Highly dissatisfied with assistance. \\
\addlinespace[3pt]
\multicolumn{5}{@{}l}{\textbf{Box lifting}} \\
\addlinespace[2pt]
6-astra & Yes & Yes & Abrupt assistance, slightly stronger than preferred. & Improved overall; abrupt, asymmetric lift onset remained. \\
6-sol & Yes & Yes & Little assistance perceived during box placement. & Improved placement support; assistance weak and delayed. \\
DeepSeek & Yes & Yes & Unwanted or jerky movements. & Improved overall; placement assistance remained unsatisfactory. \\
GLM & Yes & Yes & Unwanted or jerky movements. & Improved overall; weak lifting support, little placement support. \\
Kimi & Yes & Yes & Upward assistance continued during box placement. & Improved overall; placement assistance remained unsatisfactory. \\
\addlinespace[3pt]
\multicolumn{5}{@{}l}{\textbf{Waterbag carrying}} \\
\addlinespace[2pt]
6-astra & Yes & Yes & Support slightly too strong; assistance persisted after placement. & Improved assistance; residual support hindered arm lowering. \\
6-sol & Yes & Yes & Jerks during walking; no assistance perceived during lowering. & Less jerking; residual support hindered arm lowering. \\
DeepSeek & Yes & Yes & Assistance before grasping; arm extension impeded. & Highly dissatisfied with assistance. \\
GLM & Yes & Yes & Assistance before grasping; arm extension impeded. & Highly dissatisfied with assistance. \\
Kimi & No & Yes & Strong, persistent assistance before grasping. & Highly dissatisfied with assistance. \\
\bottomrule
\end{tabularx}
\par\vspace{3pt}
\begin{minipage}{\textwidth}
\fontsize{7.8}{9}\selectfont
Yes and No indicate whether the prescribed task was completed. Feedback entries summarize qualitative reports from one participant. After feedback denotes the policy selected after the first feedback round.
\textsuperscript{\fontsize{6}{6}\selectfont a}The 6-sol dumbbell-curl policy was retained unchanged.
\end{minipage}
\endgroup
\end{table*}

\begingroup
\widowpenalty=10000
\looseness=-1
Additional tests examined selected policies under changes in movement and loading conditions. Box-lifting tests included pausing and resuming while holding the box, small voluntary upward and downward movements, and increasing the load from 2 to 5~kg before trial onset. Dumbbell-curl policies were also tested during empty-handed curls; a separate test involved lateral arm elevation followed by elbow flexion and extension while holding a 2-kg dumbbell. Across all three tasks, additional trials involved maintaining grip and posture without actively initiating the lift. In box lifting and waterbag carrying, additional trials examined policy responses when the held object was dropped early in task execution. Policy code remained unchanged throughout these tests, and the results were not used for further policy revision. The test conditions and cues were not supplied as explicit policy inputs.\par
\endgroup


\subsection{Initial Assistance Performance}
\label{sec:initial-assistance}

\begin{figure*}[!tp]
\centering
\includegraphics[width=\textwidth]{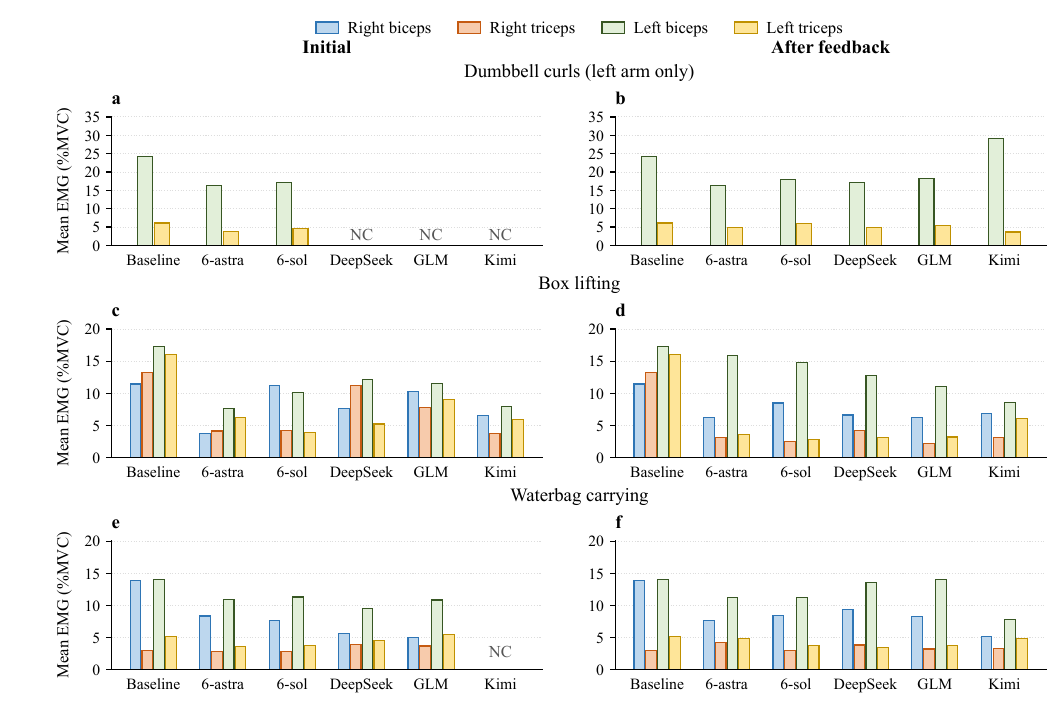}
\caption{Mean muscle activation with initial policies (left column) and policies selected after one feedback round (right column). (a,b) Dumbbell curls; (c,d) box lifting; (e,f) waterbag carrying. Bars represent task-averaged MVC-normalized EMG (\%MVC) for each muscle. The no-exoskeleton baseline is shared between the two columns for each task. NC indicates task non-completion; EMG bars are omitted for these trials. The 6-sol dumbbell-curl policy was retained after feedback.}
\label{fig:emg_results}
\end{figure*}

Figure~\ref{fig:emg_results} compares mean muscle activation under the initial policies and the policies selected after feedback. Table~\ref{tab:assistance_outcomes} summarizes task completion and participant-reported assistance for the same conditions. In the dumbbell-curl task (Fig.~\ref{fig:emg_results}(a)), mean activation of the left biceps brachii and triceps brachii with the 6-sol starting policy was 28.9\% and 23.3\% lower, respectively, than the no-exoskeleton baseline. The participant completed all three repetitions and reported a satisfactory overall experience without discomfort. With 6-astra, the corresponding reductions were 33.0\% and 39.2\%. The participant completed all three repetitions but reported assistance that felt abrupt and stronger than desired during lifting, with difficulty lowering the arm. The starting policies from DeepSeek, GLM, and Kimi also interfered with lowering, preventing completion of the prescribed three repetitions.

\begin{figure*}[!tp]
\centering
\includegraphics[width=\textwidth]{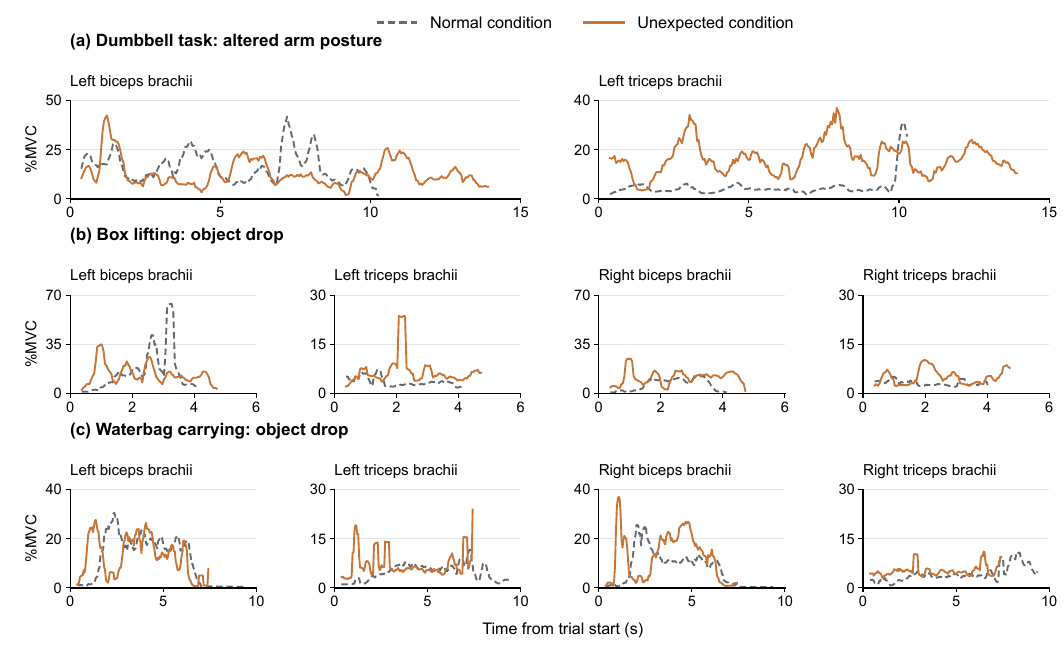}
\caption{MVC-normalized EMG during normal tasks and unexpected-condition tests using the post-feedback 6-astra policies. (a) Lateral arm elevation followed by elbow flexion and extension while holding a dumbbell. (b) Box-drop test. (c) Waterbag-drop test. Gray dashed and orange solid curves show the corresponding normal and unexpected-condition trials, respectively. Time is measured from the start of each trial; traces retain their original durations.}
\label{fig:unexpected_emg}
\end{figure*}

In box lifting, the participant completed the task with each model's starting policy. Compared with the no-exoskeleton baseline, all five starting policies showed lower mean activation in all four muscles (Fig.~\ref{fig:emg_results}(c)). With 6-astra, activation of the right and left biceps brachii was 67.8\% and 55.7\% lower, respectively, while the corresponding reductions for the triceps brachii were 68.8\% and 60.8\%. The participant reported assistance that felt abrupt and slightly stronger than preferred with 6-astra, little perceived assistance during box placement with 6-sol, unwanted or jerky movements with DeepSeek and GLM, and continued upward assistance during box placement with Kimi.

During waterbag carrying, the participant completed the task with the starting policies from 6-astra, 6-sol, DeepSeek, and GLM. Mean activation was lower than the no-exoskeleton baseline in all four muscles with 6-astra and 6-sol (Fig.~\ref{fig:emg_results}(e)). The reductions in right and left biceps brachii activation were 39.6\% and 22.0\% with 6-astra, and 44.4\% and 19.7\% with 6-sol, respectively. DeepSeek and GLM also showed lower biceps activation bilaterally, but higher right triceps activation; GLM additionally showed higher left triceps activation. The participant reported that support with 6-astra felt slightly stronger than desired and persisted after setting the bag down, whereas 6-sol produced intermittent jerks during walking and no perceived assistance during lowering. With DeepSeek and GLM, assistance began before grasping the bag and interfered with subsequent arm extension, according to participant feedback. The participant was unable to complete the task with Kimi's starting policy and reported strong, persistent assistance that began before grasping the bag.

\subsection{Feedback-Based Policy Refinement}
\label{sec:feedback-performance}

After the first round of feedback, the participant completed all three dumbbell curls with the revised 6-astra policy and reported a satisfactory experience without discomfort. Mean activation of the left biceps brachii and triceps brachii was 32.7\% and 20.6\% lower than the no-exoskeleton baseline, respectively (Fig.~\ref{fig:emg_results}(b)). The feedback analysis retained the 6-sol policy unchanged, and the participant again completed the task without reporting discomfort. The participant also completed the task with the revised DeepSeek, GLM, and Kimi policies, but reported being highly dissatisfied with the assistance provided by all three. Compared with its starting version, the revised 6-astra policy received more favorable participant feedback, with higher mean activation in both muscles. The revision reduced assistance magnitude and adjusted its onset and withdrawal to address initially excessive assistance and movement interference.

In box lifting, the participant completed the task with each policy revised after the first round of feedback. Compared with the no-exoskeleton baseline, all five revised policies showed lower mean activation in all four muscles (Fig.~\ref{fig:emg_results}(d)). With 6-astra, assistance at lift onset still felt abrupt and differed between the two sides. With 6-sol, assistance was perceptible during box placement, but felt slightly weak and delayed during both lifting and placement. The participant remained dissatisfied with the assistance provided by DeepSeek and Kimi during box placement. With GLM, assistance felt weak during lifting and was barely perceptible during placement. Compared with their starting versions, all five revised policies received more favorable overall participant feedback. Mean activation decreased in all four muscles with GLM and increased in some muscles with 6-astra, DeepSeek, and Kimi. With 6-sol, activation decreased in the right biceps and both triceps but increased in the left biceps. The revisions to 6-astra, DeepSeek, and Kimi reduced assistance magnitude to address initially excessive assistance.


During waterbag carrying, the participant completed the task with all five revised policies. Compared with the no-exoskeleton baseline, mean activation was lower in both biceps and the left triceps with all five policies, whereas right triceps activation was higher with all except 6-sol (Fig.~\ref{fig:emg_results}(f)). The reductions in right and left biceps activation were 44.6\% and 19.8\% with 6-astra, and 38.9\% and 20.4\% with 6-sol, respectively. The participant reported improved assistance with 6-astra and reduced jerking during walking with 6-sol, although residual assistance with both policies hindered arm lowering after bag placement. The participant reported being highly dissatisfied with the assistance provided by DeepSeek, GLM, and Kimi. Compared with their starting policies, 6-astra and 6-sol showed higher mean activation in some muscles. Their revisions adjusted assistance timing and withdrawal to reduce movement interference and improve smoothness.

\subsection{Responses to Changes in Task Conditions}

\label{sec:unexpected-conditions}

The participant completed box lifting with the unchanged 6-astra, DeepSeek, GLM, and Kimi policies during pauses followed by resumed motion and small voluntary upward and downward movements while holding the box. Some tests elicited continued assistance during pauses, delayed assistance during lowering, and less satisfactory assistance during box placement. With the box mass increased from 2 to 5~kg, the participant also completed the task with all four policies, although assistance felt weak. During unloaded curls, the participant completed the task with both 6-astra and 6-sol and reported assistance similar to that experienced while holding the dumbbell.

Across the three tasks, the participant grasped the object but temporarily refrained from actively initiating the lift. The participant reported no lifting of the load during this interval.

The 6-astra policy provided insufficient assistance when the participant raised the arm laterally and then flexed and extended the elbow while holding a 2-kg dumbbell, according to participant feedback (Fig.~\ref{fig:unexpected_emg}(a)). Relative to normal curls with the same post-feedback policy, whole-trial mean activation increased from 4.86\% to 16.51\% MVC in the left triceps brachii and decreased from 16.33\% to 13.13\% MVC in the left biceps brachii. The insufficient assistance in this altered posture may reflect the greater effort required to support the exoskeleton against gravity.

In the box-drop test with 6-astra, the participant reported resistance before placing the box on the table, while the hands were still supporting it (Fig.~\ref{fig:unexpected_emg}(b)). Compared with the normal trial using the same post-feedback policy, whole-trial mean activation increased from 3.56\% to 6.67\% MVC in the left triceps brachii and from 3.18\% to 4.98\% MVC in the right triceps brachii. Right biceps activation also increased, whereas left biceps activation decreased.

In the waterbag-drop test with 6-astra, the participant reported continued assistance after the drop and found the assistance unsatisfactory (Fig.~\ref{fig:unexpected_emg}(c)). Whole-trial mean activation was higher in all four recorded muscles than in the normal trial using the same post-feedback policy. Left and right triceps brachii activation increased from 4.80\% to 7.24\% MVC and from 4.22\% to 5.27\% MVC, respectively.

Overall, the 6-astra policies showed limited adaptability to altered arm posture and object-drop conditions. Compared with the corresponding normal trials, the participant reported less satisfactory assistance, characterized by insufficient support, persistent assistance after object release, and resistance during box placement.


\FloatBarrier
\section{Conclusion}
A framework for self-improving exoskeleton assistance policies through human feedback has been presented. Assistance code was autonomously generated and revised by an intelligent agent using subjective feedback, movement performance, and muscle activation measurements. The framework was evaluated across three tasks using GPT, DeepSeek, GLM, and Kimi. After one feedback round, the participant completed all three tasks with the selected policies, while perceived assistance varied across models and tasks. The selected 6-astra and 6-sol policies reduced biceps activation relative to the no-exoskeleton baseline across all three tasks. However, limited adaptability to unexpected situations was observed. Further improvements in robustness and safety will be investigated.

\end{document}